\documentclass[letterpaper]{article} 
\usepackage{aaai2026}  
\usepackage{times}  
\usepackage{helvet}  
\usepackage{courier}  
\usepackage[hyphens]{url}  
\usepackage{graphicx} 
\usepackage{natbib}  
\usepackage{caption} 
\usepackage{algorithm}
\usepackage{algorithmic}

\usepackage{newfloat}
\usepackage{listings}
\usepackage{booktabs}       
\usepackage{subfigure}
\usepackage{multirow}
\usepackage{array}
\usepackage{amsmath}

\DeclareCaptionStyle{ruled}{labelfont=normalfont,labelsep=colon,strut=off} 
\floatstyle{ruled}
\newfloat{listing}{tb}{lst}{}
\floatname{listing}{Listing}

\title{TawPipe: Topology-Aware Weight Pipeline Parallelism for Accelerating Long-Context Large Models Training}
\author{
    Houming Wu, Ling Chen\thanks{Corresponding Author}\\
}
\affiliations{
    State Key Laboratory of Blockchain and Data Security\\ College of Computer Science and Technology\\
    Zhejiang University

    \{houmingwu, lingchen\}@cs.zju.edu.cn
}

\usepackage{bibentry}

\begin{document}

\maketitle

\begin{abstract}
Training large language models (LLMs) is fundamentally constrained by limited device memory and costly inter-device communication. Although pipeline parallelism alleviates memory pressure by partitioning models across devices, it incurs activation communication overhead that scales linearly with sequence length, limiting efficiency in long-context training. Recent weight-passing approaches (e.g., WeiPipe) mitigate this by transmitting model weights instead of activations, but suffer from redundant peer-to-peer (P2P) transfers and underutilized intra-node bandwidth. We propose \textbf{TawPipe}—\textbf{\underline{t}}opology-\textbf{\underline{a}}ware \textbf{\underline{w}}eight \textbf{\underline{pipe}}line parallelism, which exploits hierarchical bandwidth in distributed clusters for improved communication efficiency. TawPipe: (i) groups devices based on topology to optimize intra-node collective and inter-node P2P communication; (ii) assigns each device a fixed shard of model weights and gradients, avoiding redundant transfers; and (iii) overlaps communication with computation to hide latency. Unlike global collective operations used in fully sharded data parallelism (FSDP), TawPipe confines most communication within node boundaries, significantly reducing cross-node traffic. Extensive experiments on up to 24 GPUs with LLaMA‑style models show that TawPipe achieves superior throughput and scalability compared to state-of-the-art baselines.
\end{abstract}

\begin{links}
    \link{Code}{https://github.com/wuhouming/TawPipe}
\end{links}

\section{Introduction}

Transformer-based large language models (LLMs) have achieved impressive success across a wide range of natural language processing (NLP) tasks. Although empirical evidence consistently shows that scaling model size and training data leads to improved performance, training such large models remains fundamentally constrained by two key challenges: (1) limited device memory that restricts model capacity, and (2) high inter-device communication overhead that hampers efficiency in distributed training.

To overcome these challenges, various parallelization techniques have been proposed, including data parallelism (DP) \cite{li2014communication, jiang2020unified}, tensor parallelism (TP) \cite{shoeybi2019megatron, wang2022tesseract}, sequence parallelism (SP) \cite{li2023sequence, wu2024loongserve}, and pipeline parallelism (PP) \cite{huang2019gpipe, guan2024advances}. Among these, PP stands out for its relatively low memory footprint and high compute utilization, especially when combined with mixed-precision training \cite{micikevicius2017mixed}, activation checkpointing \cite{chen2016training, liu2025mario}, and FlashAttention \cite{dao2022flashattention}.

Conventional PP approaches (e.g., GPipe \cite{huang2019gpipe}, DAPPLE \cite{fan2021dapple}, BitPipe \cite{wu2024bitpipe} and Zero-Bubble PP \cite{qi2024zero}) divide models into pipeline stages and exchange intermediate activations across stages. This results in communication costs that scale with sequence length $S$, micro-batch size $B$, and hidden dimension $H$, yielding per-layer messages of size $BSH$. In long-context training scenarios, where $S$ is large, activation communication becomes the dominant bottleneck, severely limiting scalability.

To mitigate this, recent work has explored weight-passing pipeline parallelism (e.g., WeiPipe \cite{lin2025weipipe}), which transmits model weights instead of activations. This strategy decouples communication volume from sequence length and batch size, offering theoretical benefits when the activation-to-weight size ratio exceeds one. However, WeiPipe faces practical inefficiencies due to its exclusive reliance on peer-to-peer (P2P) communication. Specifically: (1) it fails to exploit high-speed intra-node interconnects by ignoring bandwidth asymmetry between intra- and inter-node links, and (2) it incurs redundant data transfers and elevated memory overhead, as its ring-based communication pattern requires two full communication rounds per iteration and maintains two weight buffers per device, particularly during pipeline warm-up and cool-down phases.

To this end, we propose \textbf{\underline{t}}opology-\textbf{\underline{a}}ware \textbf{\underline{w}}eight \textbf{\underline{pipe}}line parallelism (\textbf{TawPipe}), a communication-efficient framework that exploits hierarchical bandwidth and device topology in distributed clusters. TawPipe unifies two weight-passing extremes—FSDP (\cite{rajbhandari2020zero, zhao2023pytorch}) with global collectives and WeiPipe with pure P2P exchange, thereby reducing communication overhead and maximizing bandwidth utilization. Our main contributions are summarized as follows:

\begin{itemize} 
	\item We propose a group-based weight pipeline scheduler (GWPS) that partitions devices into topology-aware groups, integrating intra-node collective communication with inter-node P2P transfers to reduce cross-node traffic and improve bandwidth efficiency.
	\item We introduce a device-bound storage (DBS) strategy that assigns each device a fixed shard of model weights and gradients, avoiding redundant transfers and lowering memory consumption.
	\item We incorporate a communication-computation overlap (CCO) mechanism that asynchronously prefetches weights during computation, effectively hiding inter-node communication latency.
	\item We evaluate TawPipe on training long-context LLaMA-style \cite{touvron2023llama} models  across up to 24 GPUs, demonstrating  state-of-the-art throughput with a balanced and modest memory footprint.
\end{itemize}

Table \ref{table-symbols} summarizes the key symbols used throughout this paper for ease of reference and analysis.

\section{Related Work}

\subsection{Data Parallelism}

Data parallelism (DP) replicates a model across multiple devices and distributes input data among them \cite{li2014communication, jiang2020unified, zhou2023abs}. While effective for models of moderate size, DP becomes infeasible when the model parameters exceed the memory capacity of a single device. FSDP (\cite{rajbhandari2020zero, zhao2023pytorch}) addresses this limitation by sharding parameters, gradients, and optimizer states across devices. However, its reliance on global collective communication introduces scalability bottlenecks, especially in bandwidth-constrained environments.

\subsection{Pipeline Parallelism}

Pipeline parallelism (PP) partitions a model into sequential stages distributed across multiple devices, with activations transmitted between stages. Synchronous PP approaches (e.g., GPipe \cite{huang2019gpipe}) enforce strict  iteration-level synchronization to ensure convergence, but incur substantial activation memory overhead by scheduling all forward passes before backward passes. To mitigate this overhead, DAPPLE \cite{fan2021dapple} adopts 1F1B scheduling, and Zero-Bubble \cite{qi2024zero} further introduces gradient decoupling. Asynchronous PP approaches (e.g., PipeDream \cite{narayanan2019pipedream}, PipeMare \cite{yang2021pipemare}, and PipeDream-2BW \cite{narayanan2021memory}) improve hardware utilization by relaxing synchronization constraints, but risk gradient staleness and convergence instability. Critically, all these approaches incur communication overheads proportional to activation size, which scales with sequence length, micro-batch size, and hidden dimension, thereby limiting efficiency in long-context model training.

\begin{table}[ht]
	\small
	\begin{center}
		\begin{tabular}{ll}
			\toprule
			$P$ & The number of devices   \\
			${\rm P}_i$ &The $i$th device in a cluster \\
			$D$ & The number of groups ($P \bmod D=0$)  \\	
			$L$ & The number of layers in neural network   \\	
			$H$ & The size of hidden dimension in Transformer   \\	
			$S$ & The sequence length in Transformer   \\	
			$B$ & Micro-batch size \\
			$N$ & The number of micro-batches in a mini-batch \\
			$A^i_j$ & Activation values of $j$th layer in $i$th micro-batch \\
			$W_j$ & Weights of $j$th layer \\
			$G_j$ & Gradients of $W_j$\\
			$M_W$ & Memory consumption for the weights of one stage\\
			$M_A$ & Memory consumption for the activations of one stage\\
			$T_C$ & Communication volume for the data of one stage\\
			\bottomrule
		\end{tabular}
	\end{center}
	\caption{Symbols used in the paper.}
	\label{table-symbols}
\end{table}

Weight-passing PP approaches (e.g., WeiPipe \cite{lin2025weipipe}) mitigate this limitation by transmitting model weights instead of activations. This design reduces communication volume to a constant level, independent of sequence length or batch size, and is theoretically advantageous when the activation-to-weight size ratio exceeds one. However, WeiPipe relies on a fixed ring-based P2P communication pattern that neglects hardware topology and computational dependencies, leading to redundant transfers and suboptimal use of high-speed intra-node interconnects (e.g., NVLink).

\subsection{Topology-Aware Communication}

Topology-aware communication improves bandwidth utilization by adapting data transfers to the underlying hardware topology. This concept is implemented in communication libraries (e.g., TACCL \cite{shah2023taccl} and HiCCL \cite{hidayetoglu2024hiccl}) and further explored in distributed training systems. BytePS \cite{jiang2020unified} employs hierarchical parameter servers to decouple intra-node and inter-node communication, while TopoOpt \cite{wang2023topoopt} co-optimizes device placement and communication scheduling based on topology modeling. Although effective in enhancing collective communication for data-parallel training, these solutions lack native support for pipeline-parallelism and weight-passing schemes.

To address these gaps, we propose TawPipe. Instead of relying solely on P2P communication, TawPipe introduces hierarchical communication scheduling into weight-passing PP, enabling efficient use of both intra- and inter-node bandwidth. In addition, it integrates topology-aware device grouping with device-bound storage to align communication with hardware topology and eliminate redundant transfers.

\section{Methodology}

\subsection{Overview}

\begin{figure*}[ht]
	\centering
	\includegraphics[width=0.78\textwidth]{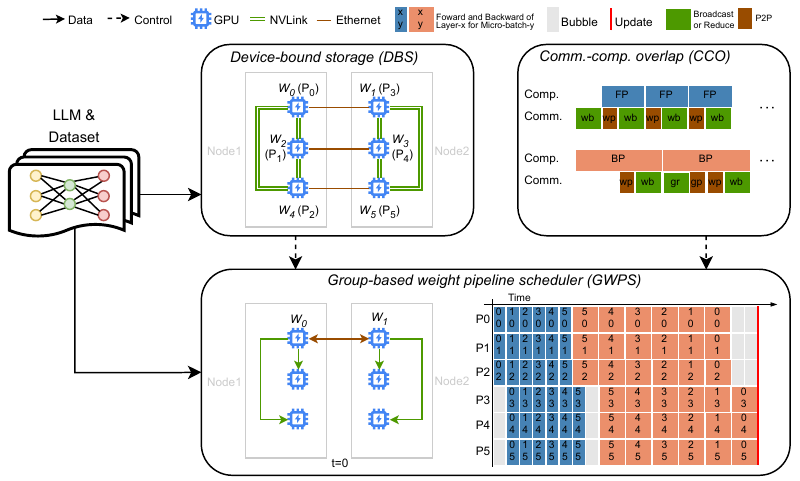}
	\caption{Overview of the TawPipe design. DBS fixes weight shards on each device, GWPS aligns communication with the underlying hardware topology, and CCO overlaps inter-group prefetching with computation. “wp”, “wb”, “gp”, and “gr” denote weight passing, weight broadcasting, gradient passing, and gradient reduction, respectively.}
	\label{fig:tawp-overview}
\end{figure*}

TawPipe comprises three tightly coupled components (as depicted in Figure \ref{fig:tawp-overview}): (i) Device-Bound Storage (DBS) assigns each device a fixed weight shard to eliminate redundant transfers and reduce memory overhead, (ii) Group-based Weight Pipeline Scheduler (GWPS) orchestrates topology-aware weight propagation to maximize intra-node bandwidth and computation efficiency, and (iii) Communication-Computation Overlap (CCO) asynchronously prefetches weights to hide inter-node communication latency. Together, these components form a unified system that minimizes communication overhead while improving throughput and scalability in distributed environments.

\subsection{Device-Bound Storage}

WeiPipe's ring-based exchange scheme requires each device to cyclically store and transmit multiple weight shards, leading to increased buffer usage and redundant data transfers. To address this, TawPipe introduces DBS, which statically assigns each layer's weights and gradients to a specific device. Communication is triggered only when a device needs to compute with a remote weight shard, ensuring that at most one shard is transferred per device at any given time.

\begin{figure}[]
	\centering
	\subfigure[Ring-based initialization]{
		\includegraphics[width=0.36\textwidth]{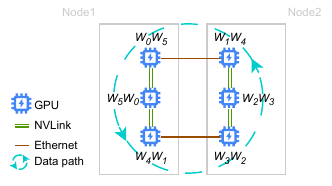}
		\label{fig:Ring-based}
	}
	\subfigure[Device-bound initialization]{
		\includegraphics[width=0.36\textwidth]{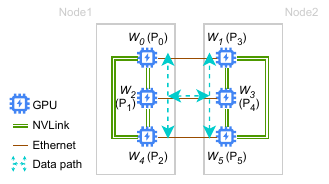}
		\label{fig:Data-binding}
	}
	\caption{Weight initialization under ring-based and device-bound strategies. Each set of three GPUs corresponds to a compute node. (a) Ring-based scheme needs to buffer and rotate two weight shards continuously. (b) Device-bound strategy stores one weight shard and initiates communication only when needed.}
	\label{fig:DBS}
\end{figure}

Figure \ref{fig:DBS} illustrates the difference in weight initialization between the two strategies for a six-GPU setup. In the ring-based strategy, each device must maintain two distinct weight buffers (e.g., $W_0$ and $W_5$ in $\rm P_0$) and complete two full communication cycles per iteration. In contrast, the device-bound strategy statically assigns a single weight shard to each device (e.g., $W_0$ to $\rm P_0$), with communication initiated only as needed. This design eliminates redundant buffer allocation and reduces communication rounds by up to 50\%. In addition, DBS integrates effectively with conventional pipeline scheduling and communication primitives (e.g., \texttt{Send/Recv} and \texttt{Broadcast/Reduce}), making it practical for scalable training for large-scale models.

\subsection{Group-Based Weight Pipeline Scheduler}

\begin{figure*}[ht]
	\centering
	\includegraphics[width=0.9\textwidth]{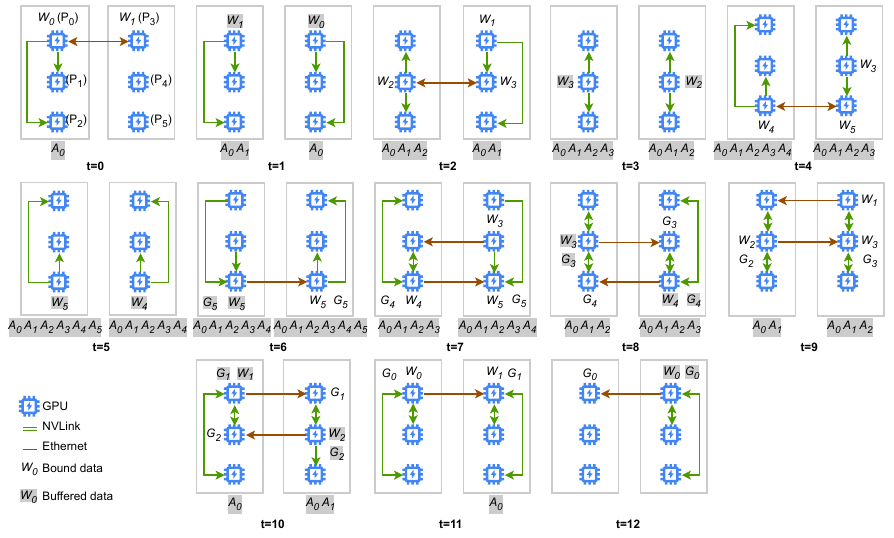}
	\caption{Group-based weight pipeline scheduling. The texts without background color represent device-bound data. The text with gray background color denotes buffered data in memory.  For instance, at $t=4$, ${\rm P}_2$ holds $W_4$, and receives $W_5$ from ${\rm P}_5$.}
	\label{fig:GWPS}
\end{figure*}

GWPS improves communication locality by organizing devices into topology-aware groups,  typically aligned with physical nodes (e.g., one group per node), and assigning model layers in an alternating pattern. Given $P$ devices and $L\bmod P=0$ layers, devices are evenly divided into $D$ groups, where $D$ generally corresponds to the number of nodes and $P$ is divisible by $D$. Specifically, group $\rm g_0$ includes devices $\{ {\rm P}_0,{\rm P}_1,...,{\rm P}_{P/D-1}\}$, group $\rm g_1$ includes $\{ {\rm P}_{P/D},{\rm P}_{P/D+1},...,{\rm P}_{2P/D-1}\}$, and so on. Each device holds exactly one weight shard. Within group ${\rm g}_k$ ($k \in [0,D{-}1]$), device ${\rm P}_{i}$ ($i \in [0,P/D{-}1]$) holds weight shard $W_{(D \cdot i + k)\bmod P}$, resulting in a balanced and interleaved layer-to-device mapping across all groups. To support efficient pipelining, each group assigns two logical roles: the master device, which initially holds the weight shard required for the current computation step, and the staging device, which asynchronously prefetches the next weight shard from a remote group for the upcoming step. 

Communication is structured hierarchically into intra- and inter-group phases. Intra-group communication exploits high-bandwidth collectives for weight broadcasting and gradient aggregation, where the master or staging device distributes weights and collects gradients within the group. Inter-group communication is restricted to lightweight P2P transfers of weights and aggregated gradients across groups. This hierarchical design localizes most traffic to intra-node links, substantially reducing cross-node communication overhead.

Training proceeds in three phases: forward, backward, and update. In the forward pass, at step $t=0$, device ${\rm P}_0$ broadcasts  weight shard $W_0$ within group $\rm g_0$, initiating parallel computation. Simultaneously, ${\rm P}_{0}$ sends $W_0$ to ${\rm P}_{P/D}$ and receives $W_1$ in return. Devices in $\rm g_0$ cache the resulting activations $A_0$ and proceed to the next layer using $W_1$, while ${\rm P}_{P/D}$ broadcasts $W_0$ within $\rm g_1$ to start computation. This staggered communication and computation pipeline proceeds until all layers are processed. The forward steps {0 to 5} are illustrated at the top of Figure~\ref{fig:GWPS}. The backward pass involves local gradient reductions within a group and inter-group transfers to the owner of the respective shard. For example, gradients for $W_{L-1}$ are first computed by devices in group $\rm g_0$, reduced to device ${\rm P}_{P/D-1}$, and then transferred to ${\rm P}_{P{-}1}$—the owner of $W_{L{-}1}$, for final aggregation and update, as illustrated in steps $6$ to $11$ (bottom of Figure \ref{fig:GWPS}). Once gradients arrive at their corresponding shard owners, updates are applied locally using colocated optimizer states, avoiding additional communication during the update phase. For example, at $t=7$, device ${\rm P}_{5}$ updates $W_{5}$ using gradient $G_{5}$ without any inter-device synchronization.

\begin{figure*}[ht]
	\centering
	\includegraphics[width=0.9\textwidth]{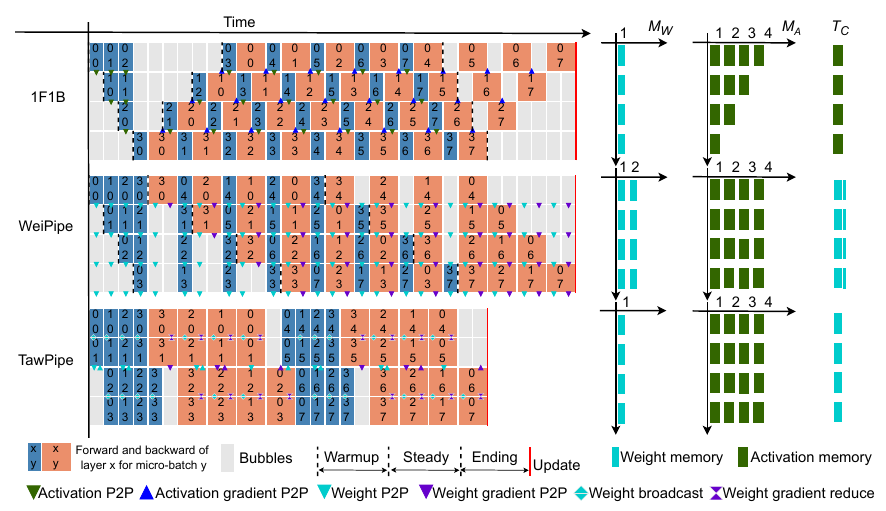}
	\caption{Comparison of TawPipe with 1F1B and WeiPipe. In the pipelining graph, the upper number in a block represents the layer index and the bottom number is the micro-batch index. TawPipe has a lower bubble ratio as the pipeline flush happens sooner in the timeline. }
	\label{fig:comparison}
\end{figure*}

By aligning the scheduling with hardware topology, GWPS minimizes cross-node communication and maximizes intra-node bandwidth utilization, leading to efficient, scalable training performance.

\subsection{Communication-Computation Overlap}

To further improve pipeline efficiency, TawPipe introduces CCO, which hides inter-group transfer latency by asynchronously prefetching remote weight shards required for the next step. Specifically, during execution at time step $t$, each staging device initiates non-blocking transfers of the weight needed for step $t{+}1$, provided it resides on another group.

CCO is implemented through dedicated memory buffers to decouple communication with computation, combined with synchronization mechanisms to ensure data consistency and avoid resource contention. For example, by leveraging non-blocking communication APIs (e.g., \texttt{torch.distributed.isend}/\texttt{irecv}), weight transfers for the next step are launched in parallel with the current forward or backward computation, effectively masking communication delays and improving pipeline utilization.

\subsection{Theoretical Analysis}

We compare TawPipe against two representative baselines: 1F1B \cite{fan2021dapple}, a widely adopted activation-passing pipeline parallelism approach, and WeiPipe \cite{lin2025weipipe}, a recent weight-passing variant. The analysis focuses on three dimensions: pipeline efficiency, memory usage, and communication volume, as illustrated in Figure \ref{fig:comparison}.

Pipeline utilization is quantified by the bubble ratio, defined as the fraction of idle time over total execution time. TawPipe achieves a lower bubble ratio of $\frac{(D-1)\cdot P+N}{(3N+D-1) \cdot P+N}$, compared to $\frac{P-1}{N+P-1}$ for both 1F1B and WeiPipe, indicating more efficient pipeline scheduling with fewer idle slots.

Memory usage is dominated by activation buffers, with weight buffers contributing to a lesser extent. Although all three approaches have comparable peak activation memory usage (approximately $BM_A$), TawPipe achieves better memory balance. In addition, through the DBS strategy, TawPipe further reduces weight buffer overhead from $2M_W$ (in WeiPipe) to $M_W$, lowering memory footprint.

Communication volume is approximated by total data received per device per iteration. For a typical LLaMA-style layer with $12H^2$ parameters and activation volume $BSH$, 1F1B transfers $2PBSH$ activations per step. In contrast, TawPipe transfers a single weight shard and its corresponding gradients per step, totaling $24H^2$. This represents a 33\% reduction compared to WeiPipe’s $36H^2$, making TawPipe particularly communication-efficient in long-sequence bandwidth-constrained settings.

\section{Experiments}
\subsection{Experimental Setup}

\begin{table*}
	\small
	\centering
	\begin{tabular}{c|c|ccc|ccc|ccc}
		\toprule
		\multirow{3}{*}{Setup}         & H       & \multicolumn{3}{c|}{1024}    & \multicolumn{3}{c|}{2048}    & \multicolumn{3}{c}{4096}    \\ 
		& S       & 4096    & 8192    & 16384   & 4096    & 8192    & 16384   & 4096    & 8192    & 16384   \\ 
		& B (*)       & 16 (4)  & 8 (1)   & 4 (1)   & 16 (4)  & 8 (1)   & 4 (1)   & 16 (4)  & 8 (1)   & 4 (1)   \\ 
		\midrule
		\multirow{6}{*}{\shortstack{Throughput \\(Tokens/\\GPU/\\second)}} &1F1B & 7212.1 & 6636.4 & 5593.7 & 3028.5 & 2886.1 & 2900.7 & \underline{1247.8} & \underline{1365.8} & 1114.2 \\
		&ZB1   & 5893.3 & 5803.1 & 5883.4 & 2884.5 & 2839.8 & 2914.8 & OOM & 1323.3 & OOM \\ 
		&ZB2   & 6021.3 & 5918.8 & 5902.8 & OOM & 2860.5 & OOM & OOM & OOM & OOM \\
		&FSDP  & 10559.2  & 8826.3  & 6750.5  & 3860.3  & 3628.4  & 3165.2  & 1128.1  & 1086.5  & 956.1  \\
		&WeiPipe  & \underline{12054.8}  & \underline{10663.1}  & \underline{8412.3}  & \underline{4008.1}  & \underline{4377.0}  & \underline{3841.7} & 1038.1  & 1158.9  & \underline{1232.4}  \\
		&TawPipe   & \textbf{13629.2}  & \textbf{11738.2}  & \textbf{8913.7}  & \textbf{5040.0}  & \textbf{4636.6}  & \textbf{4175.7}  & \textbf{1567.7}  & \textbf{1511.5}  & \textbf{1377.6}  \\ 
		\midrule
		\multirow{6}{*}{\shortstack{Memory \\(GB)}} &1F1B  & \textbf{14.5}    & \textbf{16.1}    & 20.4    & \underline{27.6}    & 28.9    & 34.8    & \underline{55.1}    & \underline{54.9}    & 62.3    \\
		&ZB1    & 34.3      & \underline{19.0}      & 36.7      & 67.6      & 35.9      & 69.3      & OOM      & 71.1      & OOM      \\
		& ZB2   & 64.5    & 34.2    &67.8    & OOM    & 66.8    & OOM    & OOM    & OOM    & OOM    \\
		&FSDP   & \underline{19.4}    & 19.4    & \textbf{19.4}    & 27.8    & \underline{27.8}     & \underline{27.8}     & \textbf{52.0}    & \textbf{52.0}    & \textbf{52.0}    \\
		&WeiPipe  & 22.0    & 22.0    & 22.0    & 29.0    & 29.0    & 29.0  & 57.8    & 57.8    & 57.8    \\
		&TawPipe   &19.6    & 19.6    & \underline{19.6}   & \textbf{27.5}    & \textbf{27.5}    & \textbf{27.5}     & 56.7    & 56.7    & \underline{56.7}   \\
		\bottomrule
	\end{tabular}
	\caption{Throughput and peak memory usage of training LLaMA-style models on 24 GPUs with NVLink and Ethernet connections. For ZB strategies, $B$ is set to 4 when $S=4096$ and $B=1$ when $S \in \{8192, 16384\}$. The best results are \textbf{bolded}. The second-best results are \underline{underlined}.}
	\label{throughput-and-mem-table}
\end{table*}

\textbf{Models and Environment.} We evaluate TawPipe using models derived from the LLaMA-2 architecture \cite{touvron2023llama} on the C4 dataset \cite{raffel2020exploring}. To assess scalability, we systematically vary the hidden dimension $H$, sequence length $S$, and number of layers $L$. All experiments are conducted on a cluster with up to 24 NVIDIA A800 GPUs (80GB each). Within each node, GPUs are interconnected via NVLink, while inter-node communication is performed over 10Gb Ethernet (10GbE).

\textbf{Baselines and Implementation.} We compare TawPipe against several state-of-the-art pipeline parallelism and memory optimization approaches:
\begin{itemize} 
	\item \textbf{1F1B} \cite{fan2021dapple}. A widely used activation-passing pipeline parallelism baseline. We use the implementation from Megatron-LM project \cite{narayanan2021efficient}. 
	\item \textbf{Zero-Bubble (ZB-1 \& ZB-2)} \cite{qi2024zero}. A recent pipeline parallelism approach that decouples weight and activation gradient computations. We use the implementation released by the authors.
	\item \textbf{FSDP} \cite{zhao2023pytorch}. An enhanced data parallelism strategy that partitions weights, gradients, and optimizer states across devices. We use the implementation based on ZeRO-3 from DeepSpeed \cite{rasley2020deepspeed}.
	\item \textbf{WeiPipe} \cite{lin2025weipipe}. A recent weight-passing pipeline parallelism approach. We use the implementation released by the authors.
\end{itemize}

TawPipe is implemented by extending WeiPipe, with modifications to the pipeline scheduler and communication engine to incorporate device-bound storage and group-based scheduling. All approaches are evaluated under identical settings, including model architecture, batch size, mixed-precision training (FP16), FlashAttention, and hardware configuration. Activation checkpointing is applied uniformly across all approaches, except for Zero-Bubble, where it offers no memory savings and incurs additional overhead \cite{lin2025weipipe}. To ensure fair comparison, all approaches use the NCCL backend \cite{jeaugey2017nccl} for communication. All reported results are averaged over multiple runs to ensure statistical robustness.

\textbf{Metrics.} We evaluate all approaches in terms of end-to-end throughput (measured in tokens per second) and peak device memory usage. In addition, we conduct ablation studies to isolate and quantify the contribution of each design component in TawPipe.

\subsection{Throughput and Memory Usage}

We evaluate the throughput and memory efficiency of TawPipe using a 48-layer transformer model across 24 GPUs. Model complexity is scaled by varying the hidden dimension $H \in \{1024, 2048, 4096\}$ and sequence length $S \in \{4096, 8192, 16384\}$, resulting in model sizes ranging from 668 million to 10 billion parameters (i.e., 668M to 10B). The global batch size is fixed at 1536, with micro-batch size $B$ adjusted according to memory constraints.

Table \ref{throughput-and-mem-table} summarizes the throughput and peak memory consumption across configurations. Key findings include:
(1) TawPipe consistently achieves the highest throughput, particularly under communication-intensive settings. For the most demanding configuration $(H, S)=(4096, 16384)$, it outperforms WeiPipe, 1F1B, and FSDP by 11.8\%, 23.6\%, and 44.1\%, respectively. (2) The performance advantage of TawPipe grows with model size. At sequence length $S=16384$, TawPipe's throughput improvement over WeiPipe increases from 6.0\% to 11.8\% as hidden dimension $H$ scales from 1024 to 4096. (3) While peak memory usage remains comparable across most approaches (excluding Zero-Bubble), TawPipe achieves better memory balance across devices and reduces weight buffer overhead through its device-bound storage strategy.

\subsection{Scalability Study}

We assess the scalability of TawPipe through both weak and strong scaling experiments. In weak scaling, we proportionally increase the number of GPUs and the global batch size to maintain a constant per-device workload. In strong scaling, we fix the global batch size while varying the number of GPUs. Considering the accommodation of a single node (i.e., 8 GPUs), all experiments use a fixed model configuration of $(S, H, L)=(16384, 1024, 48)$ with $B=2$, except for ZB-2, which uses $B=1$ due to higher memory usage.

\begin{figure}[ht]
	\centering
	\includegraphics[width=0.4\textwidth]{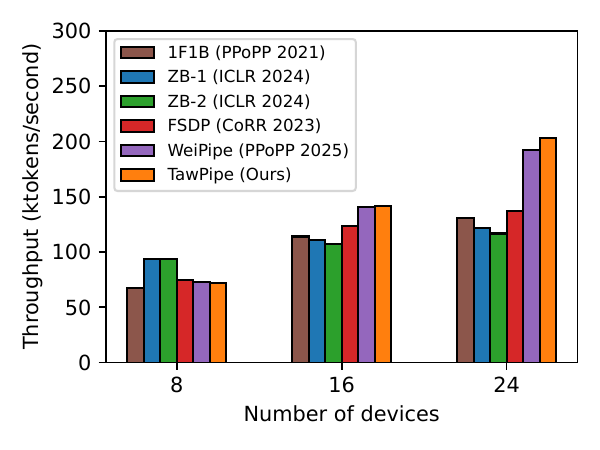}
	\caption{Weak scaling. The number of GPUs scales from 8 to 24 (8 GPUs in a node) with global batch size increasing proportionally from 512 to 1536.}
	\label{fig:weak_scaling}
\end{figure}

Figure~\ref{fig:weak_scaling} presents weak scaling results. Key observations include: (1) Conventional activation-passing approaches (i.e., 1F1B and Zero-Bubble) perform well only in tightly coupled environments (e.g., $\leq$ 8 GPUs with NVLink), where communication overhead is less impactful. (2) As the system scales across nodes, weight-passing approaches (i.e., WeiPipe and TawPipe) show greater relative speedups, with TawPipe achieving the best performance. (3) TawPipe exhibits near-linear scaling, demonstrating efficient utilization of compute and communication resources with minimal idle time.

\begin{table}[ht]
	\small
	\setlength{\tabcolsep}{1mm}
	\centering
	\begin{tabular}{lccc}
		\toprule
		Approach & NCCL-Ratio & Duration & Throughput \\
		\midrule
		1F1B (PPoPP 2021)    & 48.0\% & 105.1 & 5.59\\
		ZB-1 (ICLR 2024) & 77.6\% & 181.1 & 5.88\\
		ZB-2 (ICLR2024)& 77.5\% & 180.5 & 5.90\\
		FSDP (CoRR 2023)& \underline{33.7}\% & \underline{41.7} & 6.75\\
		WeiPipe (PPoPP 2025)& 63.7\%& 194.0 & \underline{8.41}\\
		TawPipe (Ours)& \textbf{24.1\%} & \textbf{34.7} & \textbf{8.91}\\
		\bottomrule
	\end{tabular}
	\caption{NCCL kernel time ratio, absolute duration (seconds), and throughput (kilo tokens/second). The best results are \textbf{bolded}. The second-best results are \underline{underlined}. }
	\label{tab:comm_breakdown}
\end{table}

\begin{figure}[ht]
	\centering
	\includegraphics[width=0.4\textwidth]{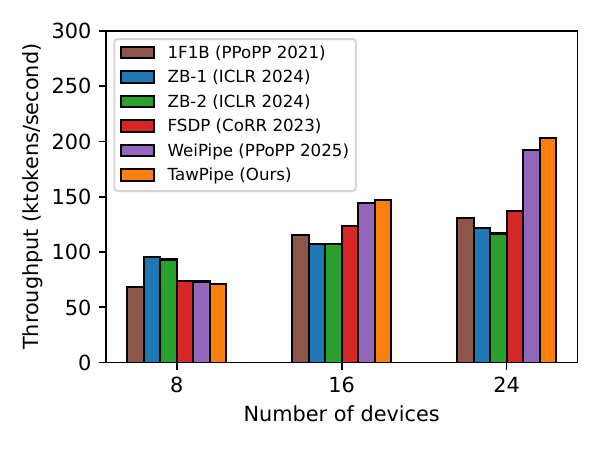}
	\caption{Strong scaling. The number of GPUs scales from 8 to 24 with the global batch size remaining 1536.}
	\label{fig:strong_scaling}
\end{figure}

Figure~\ref{fig:strong_scaling} presents strong scaling results. Key findings include: (1) Zero-Bubble strategies scale poorly due to higher memory consumption, which restricts the micro-batch size and reduces parallel efficiency. (2) Approaches relying on intensive inter-node communication (i.e., FSDP and 1F1B) suffer from degraded scalability as node number increases from 1 to 3. (3) TawPipe delivers the best strong scaling, which distributes fixed workloads efficiently without saturating communication bandwidth.

\subsection{Communication Study}

To evaluate the communication efficiency of TawPipe, we utilize NVIDIA Nsight Systems to capture GPU kernel traces and measure both the proportion of time spent on NCCL communication operations (e.g., \texttt{ncclBroadcast} and \texttt{ncclSend/Recv}), along with their absolute durations \cite{soyturk2021monitoring, shah2023taccl}.

Table~\ref{tab:comm_breakdown} presents results for a 48-layer model with $(S, H) = (16384, 1024)$ on 24 GPUs. TawPipe achieves the lowest NCCL kernel time ratio and the shortest overall communication time among all methods. It reduces NCCL execution time by up to 82.1\% compared to WeiPipe (the second best in throughput) and by 16.8\% compared to FSDP (the second best in absolute duration). These gains result from hierarchical communication that fully utilizes intra-node bandwidth and selective inter-node transfers that minimize cross-node traffic.

\begin{table}[ht]
	\small
	\setlength{\tabcolsep}{1mm}
	\centering
	\begin{tabular}{llll}
		\toprule
		Setup & 1024 & 2048& 4096\\
		\midrule
		TawPipe-w/o-GWPS & 8.59 (-3.6\%) & 3.91 (-6.5\%) & 1.26 (-8.7\%)\\
		TawPipe-w/o-CCO & 8.22 (-7.7\%) & 3.47 (-17.0\%) & 1.14 (-17.4\%)\\
		\midrule
		TawPipe & \textbf{8.91} & \textbf{4.18} & \textbf{1.38} \\
		\bottomrule
	\end{tabular}
	\caption{The performance of TawPipe and the variants (in kilo tokens/second). The best results are \textbf{bolded}. }
	\label{table-ablation}
\end{table}

\subsection{Ablation Study}

To assess the contribution of each component in TawPipe, we conduct an ablation study on two modified variants: one without the group-based weight pipeline scheduler (replaced by WeiPipe’s ring-based exchange) and another without communication–computation overlap (by disabling asynchronous prefetching).

Table \ref{table-ablation} presents the throughput results on a 48-layer model with $S=16384$ and $H \in \{1024, 2048, 4096\}$ across 24 GPUs. Key observations include: (1) TawPipe-w/o-CCO shows the largest throughput decline, confirming the effectiveness of overlapping inter-group transfers with computation. (2) TawPipe-w/o-GWPS also incurs a significant drop, underscoring the role of group-based scheduling in reducing inter-node communication and enhancing intra-node bandwidth utilization.

\section{Conclusions}

In this paper, we propose TawPipe, a topology-aware weight pipeline parallelism framework designed to accelerate the training of long-context large models. TawPipe introduces a group-based weight scheduling mechanism that exploits hierarchical hardware topology, a device-bound storage strategy that eliminates redundant data transfers, and a communication–computation overlap technique that hides inter-node latency. Experiments on LLaMA-style models across up to 24 GPUs show that TawPipe achieves state-of-the-art throughput with balanced and modest memory usage, demonstrating its scalability and efficiency for large-scale distributed training.

\bibliography{aaai2026}

@article{huang2019gpipe,
	title={{GPipe}: Efficient training of giant neural networks using pipeline parallelism},
	author={Huang, Yanping and Cheng, Youlong and Bapna, Ankur and Firat, Orhan and Chen, Dehao and Chen, Mia and Lee, HyoukJoong and Ngiam, Jiquan and Le, Quoc V and Wu, Yonghui and others},
	journal={Advances in Neural Information Processing Systems},
	volume={32},
	year={2019},
}

@inproceedings{fan2021dapple,
	title={{DAPPLE}: A pipelined data parallel approach for training large models},
	author={Fan, Shiqing and Rong, Yi and Meng, Chen and Cao, Zongyan and Wang, Siyu and Zheng, Zhen and Wu, Chuan and Long, Guoping and Yang, Jun and Xia, Lixue and others},
	booktitle={Proceedings of the ACM SIGPLAN Annual Symposium on Principles and Practice of Parallel Programming},
	pages={431--445},
	year={2021},
}

@inproceedings{narayanan2021efficient,
	title={Efficient large-scale language model training on {GPU} clusters using {Megatron-LM}},
	author={Narayanan, Deepak and Shoeybi, Mohammad and Casper, Jared and LeGresley, Patrick and Patwary, Mostofa and Korthikanti, Vijay and Vainbrand, Dmitri and Kashinkunti, Prethvi and Bernauer, Julie and Catanzaro, Bryan and others},
	booktitle={Proceedings of the International Conference for High Performance Computing, Networking, Storage and Analysis},
	pages={1--15},
	year={2021},
}

@inproceedings{narayanan2019pipedream,
	title={{PipeDream}: Generalized pipeline parallelism for {DNN} training},
	author={Narayanan, Deepak and Harlap, Aaron and Phanishayee, Amar and Seshadri, Vivek and Devanur, Nikhil R and Ganger, Gregory R and Gibbons, Phillip B and Zaharia, Matei},
	booktitle={Proceedings of the ACM Symposium on Operating Systems Principles},
	pages={1--15},
	year={2019},
}

@inproceedings{narayanan2021memory,
	title={Memory-efficient pipeline-parallel {DNN} training},
	author={Narayanan, Deepak and Phanishayee, Amar and Shi, Kaiyu and Chen, Xie and Zaharia, Matei},
	booktitle={Proceedings of the International Conference on Machine Learning},
	pages={7937--7947},
	year={2021},
}

@article{yang2021pipemare,
	title={{PipeMare}: Asynchronous pipeline parallel {DNN} training},
	author={Yang, Bowen and Zhang, Jian and Li, Jonathan and R{\'e}, Christopher and Aberger, Christopher and De Sa, Christopher},
	journal={Proceedings of Machine Learning and Systems},
	volume={3},
	pages={269--296},
	year={2021},
}

@inproceedings{rajbhandari2020zero,
	title={{ZeRO}: Memory optimizations toward training trillion parameter models},
	author={Rajbhandari, Samyam and Rasley, Jeff and Ruwase, Olatunji and He, Yuxiong},
	booktitle={Proceedings of the International Conference for High Performance Computing, Networking, Storage and Analysis},
	pages={1--16},
	year={2020},
}

@article{chen2016training,
	title={Training deep nets with sublinear memory cost},
	author={Chen, Tianqi and Xu, Bing and Zhang, Chiyuan and Guestrin, Carlos},
	journal={arXiv preprint arXiv:1604.06174},
	year={2016}
}

@inproceedings{jiang2020unified,
	title={A unified architecture for accelerating distributed {DNN} training in heterogeneous {GPU/CPU} clusters},
	author={Jiang, Yimin and Zhu, Yibo and Lan, Chang and Yi, Bairen and Cui, Yong and Guo, Chuanxiong},
	booktitle={Proceedings of the USENIX Conference on Operating Systems Design and Implementation},
	pages={463--479},
	year={2020},
}

@article{shoeybi2019megatron,
	title={{Megatron-LM}: Training multi-billion parameter language models using model parallelism},
	author={Shoeybi, Mohammad and Patwary, Mostofa and Puri, Raul and LeGresley, Patrick and Casper, Jared and Catanzaro, Bryan},
	journal={arXiv preprint arXiv:1909.08053},
	year={2019}
}

@article{li2014communication,
	title={Communication efficient distributed machine learning with the parameter server},
	author={Li, Mu and Andersen, David G and Smola, Alexander and Yu, Kai},
	journal={Advances in Neural Information Processing Systems},
	volume={27},
	year={2014}
}

@article{zhao2023pytorch,
	title={PyTorch {FSDP}: Experiences on scaling fully sharded data parallel},
	author={Zhao, Yanli and Gu, Andrew and Varma, Rohan and Luo, Liang and Huang, Chien-Chin and Xu, Min and Wright, Less and Shojanazeri, Hamid and Ott, Myle and Shleifer, Sam and others},
	journal={arXiv preprint arXiv:2304.11277},
	year={2023}
}

@article{guan2024advances,
	title={Advances of pipeline model parallelism for deep learning training: An overview},
	author={Guan, Lei and Li, Dong-Sheng and Liang, Ji-Ye and Wang, Wen-Jian and Ge, Ke-Shi and Lu, Xi-Cheng},
	journal={Journal of Computer Science and Technology},
	volume={39},
	number={3},
	pages={567--584},
	year={2024},
}

@inproceedings{qi2024zero,
	title={Zero bubble (almost) pipeline parallelism},
	author={Qi, Penghui and Wan, Xinyi and Huang, Guangxing and Lin, Min},
	booktitle={Proceedings of the International Conference on Learning Representations},
	year={2024}
}

@inproceedings{lin2025weipipe,
	title={{WeiPipe}: Weight pipeline parallelism for communication-effective long-context large model training},
	author={Lin, Junfeng and Liu, Ziming and You, Yang and Wang, Jun and Zhang, Weihao and Zhao, Rong},
	booktitle={Proceedings of the ACM SIGPLAN Annual Symposium on Principles and Practice of Parallel Programming},
	pages={225--238},
	year={2025}
}

@inproceedings{wang2022tesseract,
	title={Tesseract: Parallelize the tensor parallelism efficiently},
	author={Wang, Boxiang and Xu, Qifan and Bian, Zhengda and You, Yang},
	booktitle={Proceedings of the International Conference on Parallel Processing},
	pages={1--11},
	year={2022}
}

@inproceedings{wu2024loongserve,
	title={Loongserve: Efficiently serving long-context large language models with elastic sequence parallelism},
	author={Wu, Bingyang and Liu, Shengyu and Zhong, Yinmin and Sun, Peng and Liu, Xuanzhe and Jin, Xin},
	booktitle={Proceedings of the ACM SIGOPS Symposium on Operating Systems Principles},
	pages={640--654},
	year={2024}
}

@article{micikevicius2017mixed,
	title={Mixed precision training},
	author={Micikevicius, Paulius and Narang, Sharan and Alben, Jonah and Diamos, Gregory and Elsen, Erich and Garcia, David and Ginsburg, Boris and Houston, Michael and Kuchaiev, Oleksii and Venkatesh, Ganesh and others},
	journal={arXiv preprint arXiv:1710.03740},
	year={2017}
}

@inproceedings{liu2025mario,
	title={Mario: Near zero-cost activation checkpointing in pipeline parallelism},
	author={Liu, Weijian and Li, Mingzhen and Tan, Guangming and Jia, Weile},
	booktitle={Proceedings of the ACM SIGPLAN Annual Symposium on Principles and Practice of Parallel Programming},
	pages={197--211},
	year={2025}
}

@article{dao2022flashattention,
	title={{FlashAttention}: Fast and memory-efficient exact attention with {IO-awareness}},
	author={Dao, Tri and Fu, Dan and Ermon, Stefano and Rudra, Atri and R{\'e}, Christopher},
	journal={Advances in Neural Information Processing Systems},
	volume={35},
	pages={16344--16359},
	year={2022}
}

@article{hidayetoglu2024hiccl,
	title={{HiCCL}: A hierarchical collective communication library},
	author={Hidayetoglu, Mert and de Gonzalo, Simon Garcia and Slaughter, Elliott and Surana, Pinku and Hwu, Wen-mei and Gropp, William and Aiken, Alex},
	journal={arXiv preprint arXiv:2408.05962},
	year={2024}
}

@article{touvron2023llama,
	title={{LLaMA} 2: Open foundation and fine-tuned chat models},
	author={Touvron, Hugo and Martin, Louis and Stone, Kevin and Albert, Peter and Almahairi, Amjad and Babaei, Yasmine and Bashlykov, Nikolay and Batra, Soumya and Bhargava, Prajjwal and Bhosale, Shruti and others},
	journal={arXiv preprint arXiv:2307.09288},
	year={2023}
}

@inproceedings{soyturk2021monitoring,
	title={Monitoring collective communication among GPUs},
	author={Soyt{\"u}rk, Muhammet Abdullah and Akhtar, Palwisha and Tezcan, Erhan and Unat, Didem},
	booktitle={European Conference on Parallel Processing},
	pages={41--52},
	year={2021}
}

@inproceedings{li2023sequence,
	title={{Sequence Parallelism}: Long sequence training from system perspective},
	author={Li, Shenggui and Xue, Fuzhao and Baranwal, Chaitanya and Li, Yongbin and You, Yang},
	booktitle={Proceedings of the Annual Meeting of the Association for Computational Linguistics},
	pages={2391--2404},
	year={2023}
}

@inproceedings{rasley2020deepspeed,
	title={{DeepSpeed}: System optimizations enable training deep learning models with over 100 billion parameters},
	author={Rasley, Jeff and Rajbhandari, Samyam and Ruwase, Olatunji and He, Yuxiong},
	booktitle={Proceedings of the ACM SIGKDD International Conference on Knowledge Discovery \& Data Mining},
	pages={3505--3506},
	year={2020}
}

@inproceedings{shah2023taccl,
	title={{TACCL}: Guiding collective algorithm synthesis using communication sketches},
	author={Shah, Aashaka and Chidambaram, Vijay and Cowan, Meghan and Maleki, Saeed and Musuvathi, Madan and Mytkowicz, Todd and Nelson, Jacob and Saarikivi, Olli and Singh, Rachee},
	booktitle={USENIX Symposium on Networked Systems Design and Implementation},
	pages={593--612},
	year={2023}
}

@inproceedings{wang2023topoopt,
	title={{TopoOpt}: Co-optimizing network topology and parallelization strategy for distributed training jobs},
	author={Wang, Weiyang and Khazraee, Moein and Zhong, Zhizhen and Ghobadi, Manya and Jia, Zhihao and Mudigere, Dheevatsa and Zhang, Ying and Kewitsch, Anthony},
	booktitle={USENIX Symposium on Networked Systems Design and Implementation},
	pages={739--767},
	year={2023}
}

@inproceedings{jeaugey2017nccl,
	title={NCCL 2.0},
	author={Jeaugey, Sylvain},
	booktitle={GPU Technology Conference (GTC)},
	volume={2},
	pages={23},
	year={2017}
}

@article{raffel2020exploring,
	title={Exploring the limits of transfer learning with a unified text-to-text transformer},
	author={Raffel, Colin and Shazeer, Noam and Roberts, Adam and Lee, Katherine and Narang, Sharan and Matena, Michael and Zhou, Yanqi and Li, Wei and Liu, Peter J},
	journal={Journal of Machine Learning Research},
	volume={21},
	number={140},
	pages={1--67},
	year={2020}
}

@article{wu2024bitpipe,
	title={{BitPipe}: Bidirectional interleaved pipeline parallelism for accelerating large models training},
	author={Wu, Houming and Chen, Ling and Yu, Wenjie},
	journal={arXiv preprint arXiv:2410.19367},
	year={2024}
}

@article{zhou2023abs,
	title={{ABS-SGD}: A delayed synchronous stochastic gradient descent algorithm with adaptive batch size for heterogeneous gpu clusters},
	author={Zhou, Xin and Chen, Ling and Wu, Houming},
	journal={arXiv preprint arXiv:2308.15164},
	year={2023}
}

\end{document}